\documentclass{article}
\usepackage{spconf}
\usepackage{graphicx}
\usepackage{amsmath}
\usepackage{amssymb}
\usepackage{fixmath}
\usepackage{booktabs}
\usepackage[export]{adjustbox}
\usepackage{esint}
\usepackage{algorithm}
\usepackage{algpseudocode}
\usepackage{multirow}
\usepackage[table,xcdraw]{xcolor}
\usepackage{amsfonts}
\usepackage{gensymb}
\usepackage{xcolor}
\usepackage{graphics}
\usepackage{caption}
\usepackage{subcaption}
\usepackage{textcomp, gensymb}
\usepackage[misc]{ifsym}

\graphicspath{{./image/}}
\DeclareGraphicsExtensions{.pdf,.png,.jpg}

\usepackage[pagebackref,breaklinks,colorlinks]{hyperref}

\definecolor{red}{rgb}{0.89,0.1,0.11}
\definecolor{light_blue}{rgb}{0.36,0.54,0.66}
\definecolor{blue}{rgb}{0.0, 0.0, 1.0}

\usepackage[capitalize]{cleveref}
\crefname{section}{Sec.}{Secs.}
\Crefname{section}{Section}{Sections}
\Crefname{table}{Table}{Tables}
\crefname{table}{Tab.}{Tabs.}

\DeclareMathOperator*{\argmax}{argmax}

\newcommand{\red}[1]{{\textcolor{red}{#1}}}

\title{Adaptive Radial Projection on Fourier Magnitude Spectrum \\
for Document Image Skew Estimation}

\name{Luan Pham$^{\star}$ \qquad Phu Hao Hoang$^{\star}$ \qquad Xuan Toan Mai$^{\dagger}$  \qquad Tuan Anh Tran$^{\dagger}$}
  
\address{$^{\star}$ \small Cinnamon AI - Department of Research \\
\small Floor 10th, No. 36 Hoang Cau street, O Cho Dua Ward, Dong Da District, Hanoi City, Viet Nam.\\
$^{\dagger}$ \small Faculty of Computer Science $\&$ Engineering, Ho Chi Minh City-University of Technology (HCMUT), \\
\small 268 Ly Thuong Kiet Street, District 10, Ho Chi Minh City, Vietnam \\
$^{\dagger}$ \small Vietnam National University Ho Chi Minh City, Linh Trung Ward, Thu Duc District, Ho Chi Minh City, Vietnam. \\
\small \textit{phamquiluan@gmail.com, phuhao1998@gmail.com, mxtoan@hcmut.edu.vn, trtanh@hcmut.edu.vn $^{\dagger}$}}

\begin{document}
\ninept
\maketitle
\begin{abstract}
Skew estimation is one of the vital tasks in document processing systems, especially for scanned document images, because its performance impacts subsequent steps directly. 
Over the years, an enormous number of researches focus on this challenging problem in the rise of digitization age.
In this research, we first propose a novel skew estimation method that extracts the dominant skew angle of the given document image by applying an Adaptive Radial Projection on the 2D Discrete Fourier Magnitude spectrum.
Second, we introduce a high quality skew estimation dataset DISE-2021 to assess the performance of different estimators.
Finally, we provide comprehensive analyses that focus on multiple improvement aspects of Fourier-based methods.
Our results show that the proposed method is robust, reliable, and outperforms all compared methods.
The source code is available at \url{https://github.com/phamquiluan/jdeskew}.
\end{abstract}

\section{Introduction}
\label{sec:intro}

In a document processing system, the input image should be ensured to be in a straight position since a graceless skew angle might seriously affect the performance of subsequent steps \cite{anh2017}. Spatial information is still a crucial feature for most deep learning architectures to achieve good performance in various document processing tasks, such as document layout analysis, information extraction, and others. A skew estimation algorithm receives a digital document image and outputs the main skew angle. It is usually coupled with an affine transformation to correct the image.

\begin{figure}[t]
\centering
\includegraphics[width=0.3\textwidth]{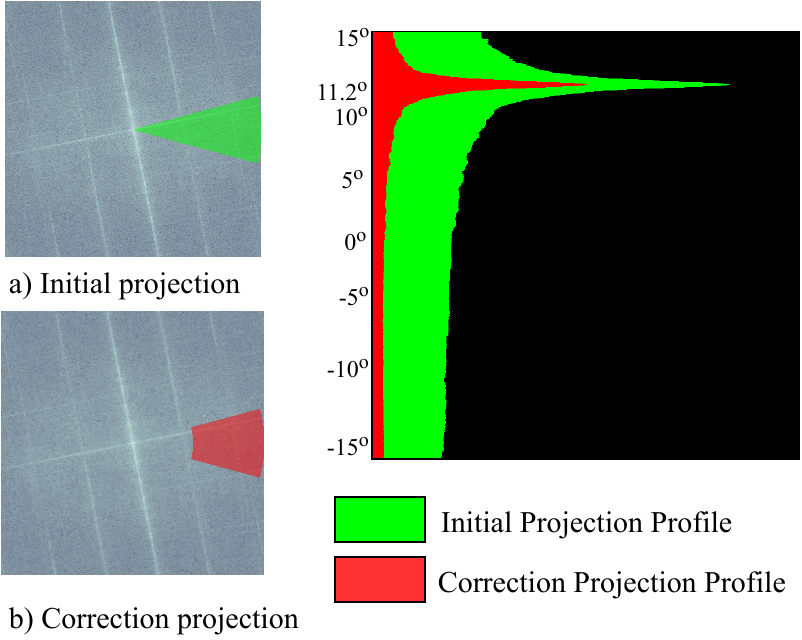}	
\caption{The overview of our adaptive projection.}
\label{fig:main_flow}
\end{figure}

By 2010, a lot of research had been conducted to tackle this problem, using different schools of thought such as Fourier transform \cite{peake1997general,lowther2002accurate}, projection profiles analysis \cite{ciardiello1988experimental,bagdanov1997projection}, Hough transforms \cite{srihari1989analysis,hinds1990document,le1994automated}, and others \cite{dhandra2006skew,smith1995simple,chen1995automatic}. 
Since the dawn of DISEC2013 competition \cite{papandreou2013icdar}, many studies \cite{fabrizio2014precise, harley2015icdar,stahlberg2015document} have been continuously carried out to improve the performance of the skew estimation task.
The angle range from $-15^{\circ}$ to $+15^{\circ}$ and the threshold of $0.1^{\circ}$ to count the number of correct estimations have been used to compare different estimators.

In this work, we propose a skew estimation method that works robustly over a larger angle range from $-44.9^{\circ}$ to $44.9^{\circ}$ degrees. We take advantage of the Fourier-based approach \cite{lowther2002accurate,peake1997general,fabrizio2014precise} because it can deal with a variety of document images without many assumptions about its kind. 
The only assumption is that the main skew angle is represented on the dominant line of the Fourier Magnitude Spectrum, see \cref{fig:main_flow}.
To extract that angle precisely, we introduce a novel Adaptive Radial Projection step.
This step is inspired by radial integration in \cite{peake1997general,postl1986detection,postl1988method}, and Radon Transform in \cite{lowther2002accurate}.
The difference is that we perform radial projection twice, then aggregate their results.
The first projection is applied directly on the magnitude from the origin as in \cite{postl1986detection,postl1988method}.
The second projection aims to produce more accurate results by discarding the DC component and low frequencies.
The empirical results show our proposed method can extract the skew angle efficiently and its performance surpasses all the compared methods \cite{pypideskew,freddeskew,koo2016robust,cmcdeskew,fabrizio2014precise}.

We also introduce a new dataset called DISE 2021 and propose a way to validate the straightness via the verification mask.
Our dataset is aggregated from three different document datasets to ensure its variety: DISEC 2013 \cite{papandreou2013icdar}, RDCL 2017 \cite{clausner2017icdar2017}, and RVL-CDIP \cite{harley2015icdar} .
We release the DISE 2021 dataset with its straight images, verification masks, and two skew versions of $15$ degrees and $44.9$ degrees.
Previously, we lacked a standard skew dataset in the range of $44.9$ degrees and the straightness verification process is still ambiguous in the literature.
Our dataset has high quality and is reviewed rigorously by our human annotators.

Additionally, we provide thorough experiments to compare the effectiveness of different factors on the Fourier-based approach under a standard dataset and criteria.
These variants are inspired by \cite{lowther2002accurate,peake1997general,fabrizio2014precise, anh2016}.
To the best of our knowledge, there are no comprehensive analyses on different improvement aspects of this approach in the literature currently. \\



\section{Proposed Method} \label{sec:method}
\subsection{Overview}
The skew estimation method we present in this paper consists of three main steps: preprocessing, 2D Discrete Fourier Transform (2D-DFT), and Adaptive Radial Projection.
First, the input image $I\in \mathbb{R}^{H\times W\times C}$, where $H,W,C$ denote its height, width and channel respectively, is subject to essential preprocessing steps that produce a binary image $B\in \{0, 1\}^{H\times W}$.
Second, the image $B$ is passed through a 2D-DFT and normalization steps to get its magnitude spectrum $M\in \mathbb{R}^{H\times W}$.
Now the oblique angle is clearly shown on the magnitude spectrum.
Finally, we apply the Adaptive Radial Projection on the spectrum $M$ to extract the output angle $\theta$ precisely.

\subsection{Adaptive Radial Projection} \label{method:radial-projection}

\begin{figure}
\begin{subfigure}[b]{0.22\textwidth}
\includegraphics[width=\textwidth]{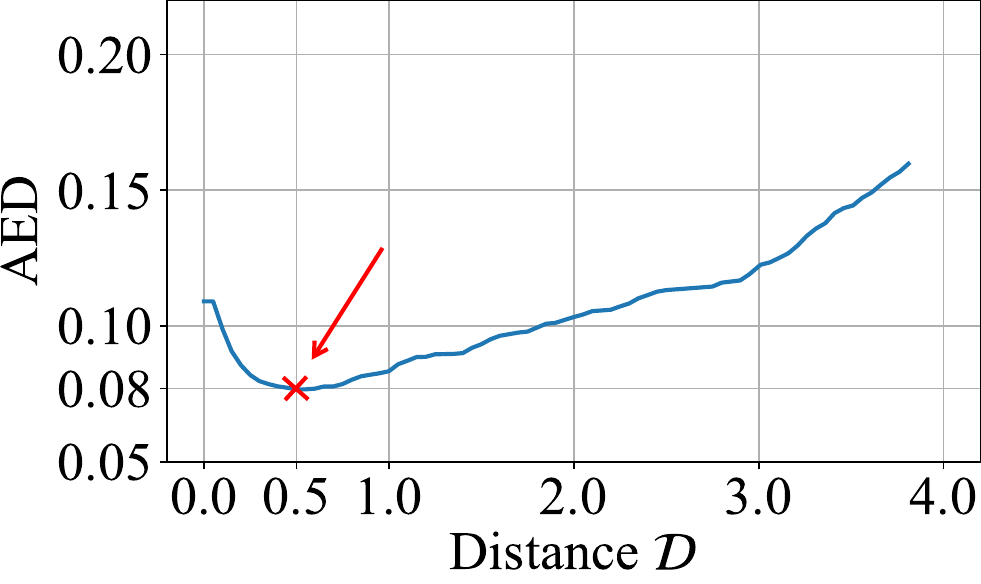}
\end{subfigure}
\hfill
\begin{subfigure}[b]{0.22\textwidth}
\includegraphics[width=\textwidth]{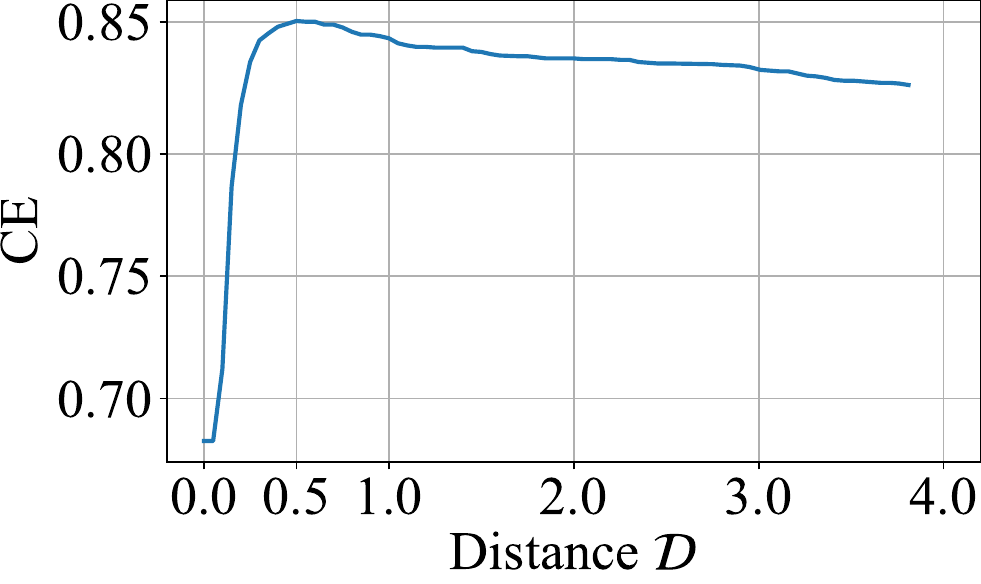}
\end{subfigure}
\caption{The change of AED and CE levels when tuning the distance $\mathcal{D}$. The chosen threshold (\red{\textbf{\texttt{x}}}) should bring the lowest AED value.} \label{fig:cor-aed-ce-d}
\end{figure}

This step is composed of two separate radial projections: the initial projection and the correction projection.
The initial projection is applied directly on the magnitude from the magnitude center as in \cite{postl1986detection,postl1988method}.
The correction projection is inspired by \cite{peake1997general,lowther2002accurate}.
However, instead of applying a square-shaped mask \cite{peake1997general} or a donut-shaped mask \cite{lowther2002accurate}, we simply move the left-most element of the integration line away from the magnitude center a distance $\mathcal{W}$, see \cref{fig:main_flow}b.
Finally, we aggregate these two outputs in a way that leverages the advantages and mitigates the disadvantages of the second projection as we analyse in \cref{ablation:discarding-dc-component-and-small-frequencies} via ablation experiments on discarding the DC component and low spatial frequencies.

Given the spectrum $M\in \mathbb{R}^{H\times W}$, the projection radius $\mathcal{R} = \min(H,W)$, and the angle list $[\theta_{\min}, \dots, \theta_{\max}]$, the initial projection value $\mathcal{A}(\theta_i)$ at the angle $\theta_i$ is calculated as follows, see \cref{fig:main_flow}a
\begin{equation}
\mathcal{A}(\theta_i) = \sum_{s=0}^\mathcal{R} M[c_y + s\cdot \cos(\theta_i), c_x - s \cdot \sin(\theta_i)],
\end{equation}

where ($c_x, c_y$) is the center coordinate, and $M[y,x]$ is the value of magnitude spectrum at coordinate $(x,y)$ 
The correction projection value $\mathcal{B}(\theta_j)$ at angle $\theta_j$ is computed as follows, see \cref{fig:main_flow}b.
\begin{equation}
\mathcal{B}(\theta_j) = \sum_{s=\mathcal{W}}^\mathcal{R} M[c_y + s \cdot \cos(\theta_j), c_x - s \cdot \sin(\theta_j)].
\end{equation}
The candidate angles $\theta_a$ and $\theta_b$ is determined by
\begin{equation}
\theta_a = \argmax_{m}\mathcal{A}(m), \hspace{1mm} \theta_b = \argmax_{n}\mathcal{B}(n)
\end{equation}
The final output angle $\theta_\mathcal{F}$ is aggregated by the following rule:
\begin{equation} \label{eq:choose-theta}
	\theta_{\mathcal{F}}= \begin{cases} \theta_a, \text{if}\ | \theta_a - \theta_b | > \mathcal{D} \\ \theta_b,  \text{otherwise} \end{cases}.
\end{equation}

\noindent
\textbf{The search of parameters.}
The window size $\mathcal{W}$ and distance $\mathcal{D}$ are chosen as follows. 
Firstly, we use $\theta_{\mathcal{B}}$ as the default output and search for $\mathcal{W}$ which brings the highest correct estimation rate \cite{papandreou2013icdar}, on the development set.
Due to the computational expensive, the coarse search space has 20 elements from 0 to $\frac{1}{3} \times H$.
The fine space includes another 20 elements around the coarse result with a deviation of $\frac{2}{30}\times H$.
The search space is configured empirically, which could be improved.
Secondly, we find $\mathcal{D}$, which brings the optimal Average Error Deviation score on our system.
In \cref{fig:cor-aed-ce-d}, at the left-most, $\mathcal{D} = 1$, which means $\theta_{\mathcal{A}}$ is the output angle by default.
At the right corner, $\theta_{\mathcal{B}}$ is the output angle without consideration. Within that range, the output angle is chosen as in \cref{eq:choose-theta}. The suitable $\mathcal{D}$ minimizes the AED score while possibly bringing the highest CE. The detailed configuration is presented in \cref{sec:thresh-config}.

\section{Data}
\label{sec:exp-data}
\subsection{Verification Mask} \label{sec:verification-mask}
\begin{figure}
\begin{subfigure}[b]{0.22\textwidth}
\centering
\includegraphics[width=\textwidth]{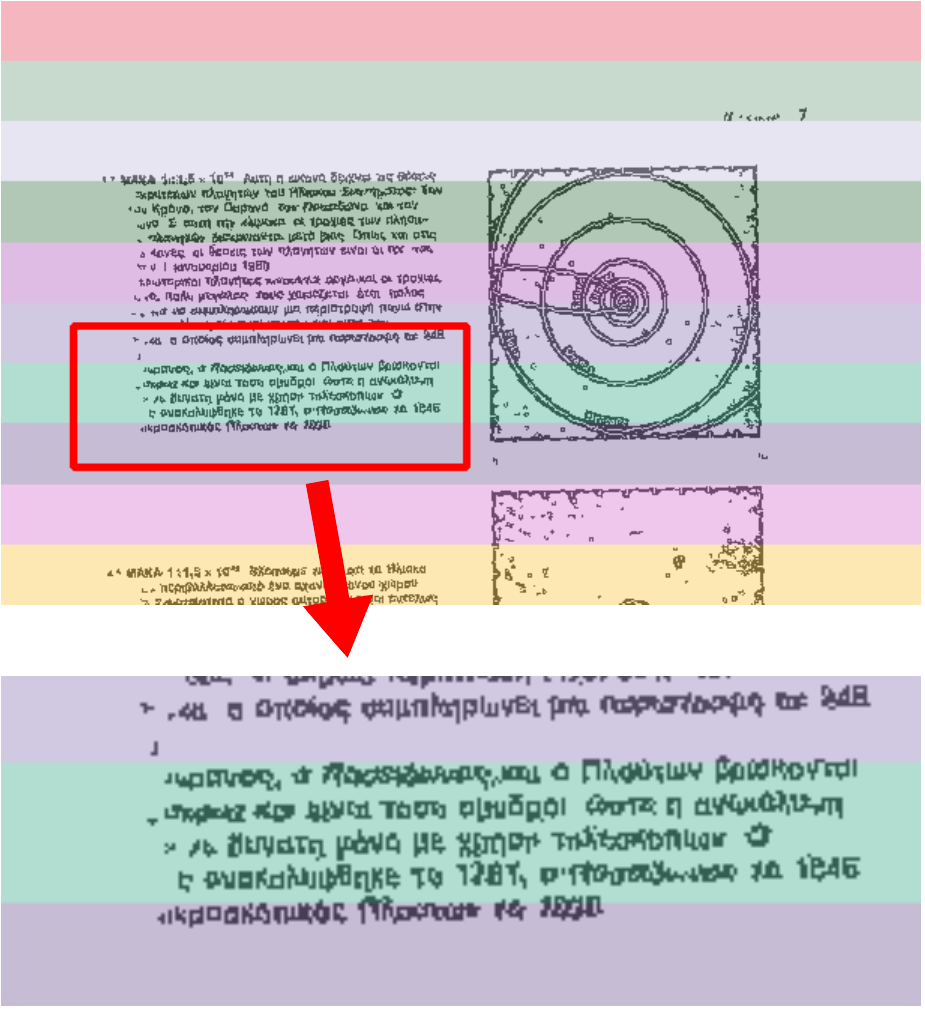}
\end{subfigure}
\hfill
\begin{subfigure}[b]{0.22\textwidth}
\centering
\includegraphics[width=\textwidth]{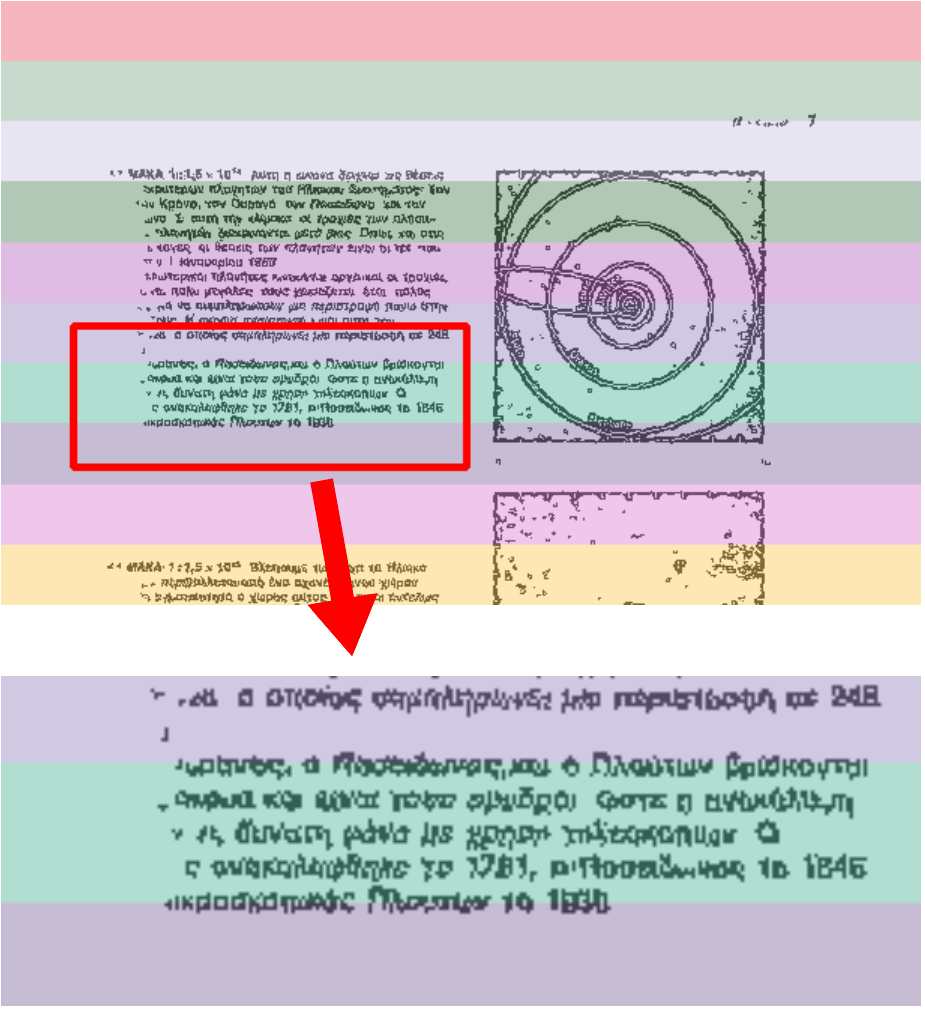}
\end{subfigure}
\caption{An example of low-quality annotation in DISEC2013, deskewed by DISEC2013 ground-truth angle (left) and  our system (right).}
\label{fig:disec-wrong-label}
\end{figure}

Although \cite{papandreou2013icdar} and \cite{hull1998document} argued that the skew angle of $0.1^{\circ}$ might be detectable to a human observer, we see that such a small angle is hard to verify normally and it might lead to unexpected biases \cite{stahlberg2015document}. We propose the  verification mask to ensure the image is in a straight position during the data review process. Firstly, the images are overlaid with a color mask horizontally or vertically. Secondly, the annotators will draw three to five red boxes that clearly show the alignment of different document components such as text lines, ruling lines, tables, and figures. In this way, we can detect noisy samples easily, see \cref{fig:disec-wrong-label}.

\subsection{DISE 2021}

In this research, we build a new DISE2021 dataset from $95$ images from DISEC2013 dataset \cite{papandreou2013icdar}, $70$ images from RDCL dataset \cite{clausner2017icdar2017}, and $324$ images from RVL-CDIP dataset \cite{harley2015icdar}.
The composed dataset contains various types of documents, multiple languages, and typography features. Firstly, all the images are ensured and verified to be in a straight position, as described in \cref{sec:verification-mask}.
Secondly, we use the same generating algorithm as in \cite{papandreou2013icdar} to generate skew images in the range $-15^{\circ}$ to $+15^{\circ}$.
The dataset is split into two development/test sets by a ratio of $0.7/0.3$ that results in 3399 development images and 1491 testing images.
When generating the skew dataset in the range from $-44.9^{\circ}$ to $44.9^{\circ}$, we double the augmented image that results in $6980$ development images and $2800$ testing images. From now on, without clearly specifying DISE 2021 ($44.9^{\circ}$), the dataset of $15^{\circ}$ degrees is used to compare the performance in experiments by default.

At the time of conducting this research, we are unable to obtain the official version of DISEC 2013 dataset \footnote{\url{http://www.iit.demokritos.gr/~alexpap/DISEC13/icdar2013_benchmarking_dataset.rar}}.
The version we use in this research is obtained by contacting \cite{bezmaternykh2020document}. From manual inspection, we found that there are some samples, which do not fit to our context, where the text orientation is more considerate, see \cref{fig:disec-wrong-label}. In 2015, \cite{stahlberg2015document} also figured out a similar problem when they got a higher Correct Estimation score by simply subtracting their output with a small offset. From these points, we see that tuning parameters for higher CE in this dataset might lead to wrong conclusions.

\section{Experiment}
\label{sec:exp}
\subsection{Criteria}
To evaluate the robustness of skew estimators, we use three standard criteria AED, TOP80, and CE \cite{papandreou2013icdar}.
Nevertheless, we also inspect the sorted absolute error curve to supply a meaningful insight of the error distribution and to observe the worst estimation.

\subsection{Threshold Configurations} \label{sec:thresh-config}

\noindent
\textbf{Image height}. Because the size of the document image varies remarkably from a small letter to a big poster or drawing document, it affects the system speed as well as the threshold design. To address this problem, we resize the input image to a pre-defined image height while preserving its aspect ratio. We use five different heights in this research $1024, 1500, 2048, 3072, 4096$ for performance comparison in terms of both accuracy and speed.

\noindent
\textbf{The range of $\mathbold{\theta}$}. Since the projection range of our Adaptive Radial Projection is configurable and the angle range is known beforehand, we set the couple $\{\theta_{\min}, \theta_{\max}\}$ equal to $\{-15, 15\}$ and $\{-44.9, 44.9\}$ when dealing with the corresponding range.

\noindent
\textbf{The window size $\mathcal{W}$ and the distance $\mathcal{D}$}.
After searching, we have 5 sets of parameters $\{H, \mathcal{W}, \mathcal{D}\}$ for each image height config.
$\{1024, 247, 0.7\}$, $\{1500, 328, 0.55\}$, $\{2048, 304, 0.55\}$, $\{3072, 307, 0.45\}$, $\{4096, 250, 0.5\}$.

\subsection{Analysis of the impact of different factors}

In this section, we aim to verify the performance gain potential of the following ideas:
(\textit{a}) Split the image by $N\times N$ parts.
(\textit{b}) Which is better, between Power spectrum and Magnitude spectrum?
(\textit{c}) Discarding the DC component and low spatial frequencies.

For simplicity, we use the baseline estimator as described in \cref{sec:method} except that it uses $\theta_\mathcal{A}$ as default without applying the correction projection result $\theta_\mathcal{B}$ to realize experiments in this section. 

\subsubsection{Image division} \label{ablation:image-division}

This experiment aims to confirm the impact of image division. 
It was proposed by \cite{peake1997general} and \cite{lowther2002accurate}. 
The general idea is instead of directly examining the Magnitude Spectrum of the whole image; we do it on image blocks, then find a way to aggregate their output.
In this experiment, we test this idea with multiple block sizes, which are proportional to the image height $H$. Firstly, we tile the input image to multiple parts of block size $N\times N$ and convert each part to the frequencies domain. Secondly, we normalize the magnitude spectrum of each block and average them. Finally, we use radial projection analysis to extract the angle from the combined magnitude spectrum.
\Cref{table:res-image-division-effect} indicates that the level of efficiency steadily decreases as we split the image into smaller parts.
For instance, the CE scores drop from $0.47$ to $0.01$ and from $0.62$ to $0.03$ in the image heights of $1024$ and $2048$ respectively.
We see that there are two main reasons for the deficiency: the outlier blocks and the smallness of magnitude spectrum affect the overall result.

Obviously, the CE only counts if and only if $E(i)\leq0.1$ \cite{papandreou2013icdar}.
Geometrically, $\frac{x}{y} = \tan(\theta)$, so to show 1 pixel in $x$, we need $\frac{1}{\tan(0.1)} \approx 573$ pixels in $y$.
It means the square image must have the size of $573 \times 2 = 1146$ pixels.
This not only explains the plummet but also the climb of performance when we increase the image height.

\begin{table}
\centering
\caption{Effects of image division with different block size.} \label{table:res-image-division-effect}

\resizebox{0.85\linewidth}{!}{
\begin{tabular}{c|ccc|ccc}
\hline
\multirow{2}{*}{Block Size} & \multicolumn{3}{c|}{1024} & \multicolumn{3}{c}{2048} \\ \cline{2-7} 
 & AED & TOP80 & CE & AED & TOP80 & CE \\ \hline \hline
$N=h * 0.1$ & 0.95 & 0.75 & 0.01 & 0.91 & 0.55 & 0.03 \\
$N=h * 0.2$ & 0.56 & 0.43 & 0.03 & 0.51 & 0.33 & 0.02 \\
\ldots & \ldots & \ldots & \ldots & \ldots & \ldots & \ldots \\
$N=h * 0.9$ & 0.16 & 0.11 & 0.39 & 0.15 & 0.08 & 0.57 \\
$N=h$ & 0.14 & 0.09 & \textbf{0.47} & 0.12 & 0.07 & \textbf{0.62} \\ \hline
\end{tabular}
}
\end{table}

\subsubsection{Power spectrum and magnitude spectrum} \label{ablation:power-spectrum-and-magnitude-spectrum}

While \cite{fabrizio2014precise} leverages Magnitude spectrum $M$, \cite{postl1986detection,postl1988method} use Power Spectrum $\mathcal{P} $, where $\mathcal{P}(u,v) = M(u,v)^2$. In this work, we also conduct the quantitative comparison of these twos when using the radial projection to extract the skew angle from the spectrum. In \cref{table:res-magnitude-power-spectrum}, it can be clearly seen the magnitude spectrum can reach remarkably higher performance in terms of evaluation score. While the AED of magnitude spectrum trials around $0.12$ to $0.14$, the power spectrum results fluctuate from $0.19$ to $0.2$. The Correct Estimation rate between these two experiments also has a large margin. In experiments of image height $1024$ and $2048$, the magnitude spectrum gets $47\%$ and $62\%$ correct rate while the opponent gets only $28\%$ and $35\%$ accordingly.

\begin{table}
\centering
\caption{Comparison of Magnitude and Power Spectrum} \label{table:res-magnitude-power-spectrum}
\resizebox{0.9\linewidth}{!}{
\begin{tabular}{l|ccc|ccc}
\hline
\multirow{2}{*}{} & \multicolumn{3}{c|}{1024} & \multicolumn{3}{c}{2048} \\ \cline{2-7} 
 & AED & TOP80 & CE & AED & TOP80 & CE \\ \hline \hline
Magnitude Spectrum & 0.14 & 0.09 & \textbf{0.47} & 0.12 & 0.07 & \textbf{0.62} \\
Power Spectrum & 0.2 & 0.14 & 0.28 & 0.19 & 0.15 & 0.35 \\ \hline
\end{tabular}
}
\end{table}

%

\subsubsection{Discarding the DC component and small frequencies} \label{ablation:discarding-dc-component-and-small-frequencies}

The effectiveness of removing the DC component and small frequencies in the Fourier spectrum is investigated in this experiment. Previously, \cite{peake1997general} uses a squared zero window, while \cite{lowther2002accurate} uses a round-shaped zero mask to override the spectrum center. In this work, we simply move the start point of the radial projection far from the spectrum center a distance $\mathcal{W}$, see \cref{fig:main_flow}b.
The results of the experiment are shown in \cref{table:res-discard-dc-and-small-frequencies}. On the one hand, significant improvement in the correct estimation rate of all image heights is clearly shown. There is an increase from $47\%$ to $65\%$ percent in the image height of $1024, 58$ to $74$, and $62$ to $78$ in the image heights of $1500$ and $2048$ respectively.
We conclude that it is possible to reach a surprisingly high correct estimation rate by having a right tuned $\mathcal{W}$.

On the other hand, it is noticeable that while we are expecting the AED to maintain a negative correlation with CE, which means higher CE should couple with lower AED, but it turns out that this is not true in this experiment. The average error goes up with the correct estimation rate. This phenomenon shows over $10^{\circ}$ worst estimation at the tail of the sorted error curve range from $5^{\circ}$ to $15^{\circ}$, causing the AED to go up dramatically.
In case of image height 3072, even though the best 80 percent of estimation results are able to achieve $0.04$ in TOP80 score, the total average error is up to a surprisingly high number of $0.74$.
Inspecting the spectrum magnitude, the zero mask in our expectation could help to remove the noisy signals, but in bad cases, it might also remove the dominant "line" in the magnitude spectrum. Hence, this causes difficulties for the later radial projection to get the right results. From the industrial point of view, this system might ruin the user experience when it gives completely wrong estimation.

To increase in CE without bringing the negative effect to AED, we perform radial projection twice as described in \cref{method:radial-projection},
the second angle $\theta_{\mathcal{B}}$ is chosen if it does not differ from $\theta_{\mathcal{A}}$ greater than a distance $\mathcal{D}$.
By this way, we gain improvements not only in the significant reduction in AED but also in the increase in CE rate.
For example, the best AED goes from $0.74$ down to one-tenth and the CE increases from $0.81$ to $0.86$ in the image height of 3072 pixels, see Table \cref{table:res-our}.

\begin{table}
\centering
\caption{Efficiency comparison when discarding the DC component and small frequencies by different window size on our baseline.} \label{table:res-discard-dc-and-small-frequencies}
\resizebox{0.8\linewidth}{!}{
\begin{tabular}{c|ccc|ccc}
\hline
\multirow{2}{*}{Window Size} & \multicolumn{3}{c|}{1024} & \multicolumn{3}{c}{3072} \\ \cline{2-7} 
 & AED & TOP80 & CE & AED & TOP80 & CE \\ \hline
$W=15$ & 0.14 & 0.09 & 0.47 & 0.1 & 0.06 & 0.68 \\
$W=35$ & 0.13 & 0.09 & 0.49 & 0.1 & 0.06 & 0.69 \\
$W=55$ & 0.13 & 0.09 & 0.51 & 0.11 & 0.06 & 0.7 \\
$W=75$ & 0.13 & 0.09 & 0.53 & 0.11 & 0.06 & 0.72 \\
 &  &  &  &  &  &  \\
$W=h/10$ & 0.13 & 0.08 & 0.55 & 0.74 & 0.04 & \textbf{0.81} \\
$W=h/8$ & 0.14 & 0.08 & 0.57 & 1.23 & 0.04 & 0.77 \\
$W=h/6$ & 0.14 & 0.07 & 0.6 & 2.19 & 0.36 & 0.65 \\
$W=h/4$ & 0.18 & 0.07 & \textbf{0.65} & 5.03 & 2.9 & 0.32 \\ \hline
\end{tabular}
}
\end{table}

\subsection{Experimental Results} \label{res:proposed-method}

\begin{figure}
\centering
\begin{subfigure}[b]{0.23\textwidth}
\includegraphics[width=\textwidth]{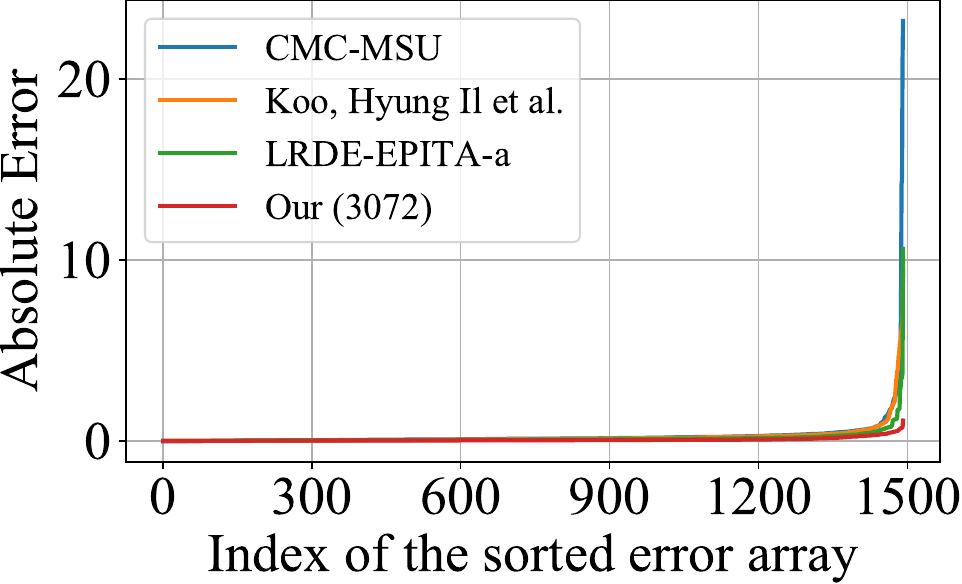}
\caption[center]{}
\end{subfigure}
\hfill
\begin{subfigure}[b]{0.23\textwidth}
\includegraphics[width=\textwidth]{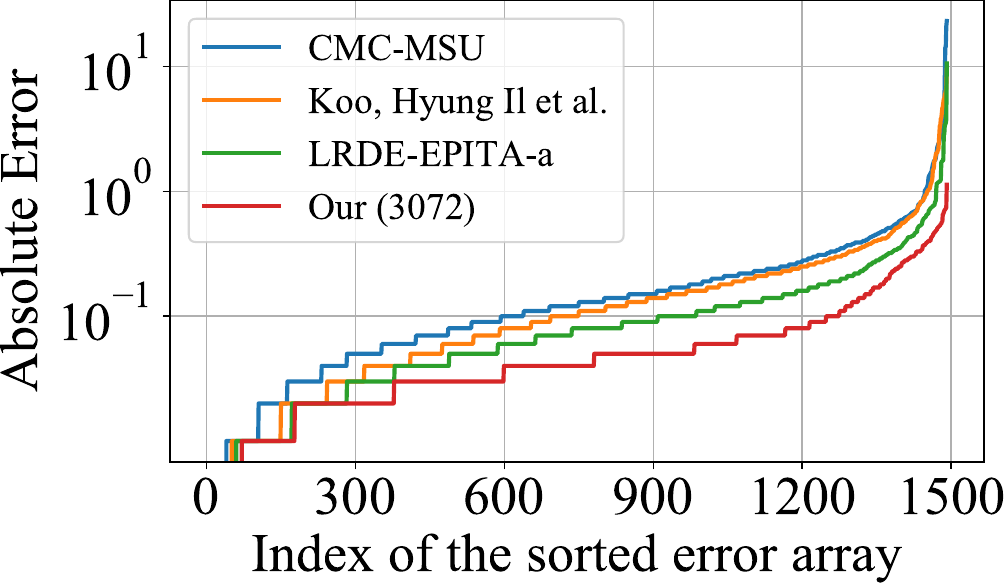}
\caption[center]{} \label{fig:experimental-result-error-curve-log}
\end{subfigure}
\caption[center]{(a) The sorted absolute error curve and (b) its log-scale version of four different methods on DISE 2021 ($15^{\circ}$) dataset.} \label{fig:experimental-result-err-curve}
\end{figure}

\Cref{table:res-our} illustrates the evaluation scores of different methods on all three datasets: DISEC 2013, DISE 2021 ($15^{\circ}$), and DISE 2021 ($49.9^{\circ}$). The proposed method outperforms all other methods on two DISE 2021 datasets.
The configuration of  $\{H,\mathcal{W},\mathcal{D}\}=\{3072, 307, 0.45\}$ brings the best scores of $0.07, 0.04, 0.86$ on the DISE 2021 ($15^{\circ}$) dataset and $0.06, 0.02, 0.88$ on the DISE 2021 ($44.9^{\circ}$) dataset in the AED, TOP80, and CE criteria respectively. While the current solutions of CMC-MSU \cite{cmcdeskew,papandreou2013icdar} and LRDE-EPITA-a \cite{fabrizio2014precise,papandreou2013icdar} do not support the range of $44.9^{\circ}$, \cite{koo2016robust} shows competitive performance on this range.
\cite{pypideskew} and \cite{freddeskew} show a reasonable performance on the DISE 2021 ($15^{\circ}$) dataset when they reach 0.24 and 0.09 respectively in the TOP80 criterion, but their AED scores are too high compared to others.

In the matter of the Worst Error (WE), our method maintains a low level around $1^{\circ}$ over all configurations. While the other approaches go from around $10^{\circ}$ to over $20^{\circ}$, the implementation of PypiDeskew and FredsDeskew have the Worst Error over 100 degrees. In addition, figure \cref{fig:experimental-result-err-curve} presents the sorted absolute error curve of four different methods. By examining this curve, we see that all four methods perform excellently over $95\%$ of cases. The difference reveals in the last $5\%$ of worst cases. The log-scale version of the error curve, in Figure \cref{fig:experimental-result-error-curve-log}, shows that our method consistently archieve better compared to the other three solutions. This insight is worthy to verify the reliability when comparing different methods on a skew estimation dataset, especially when handling hard cases. 

\begin{table}
\centering
\caption{Quantitative comparison of different methods on the DISE2021 $15^{\circ}$ and $44.9^{\circ}$ datasets. \label{table:res-our}}
\resizebox{\linewidth}{!}{
\begin{tabular}{c|cccc|cccc}
\hline
\multirow{2}{*}{Method} & \multicolumn{4}{c|}{DISE 2021} & \multicolumn{4}{c}{DISE 2021 (45)} \\ \cline{2-9} 
 & \multicolumn{1}{c|}{AED} & \multicolumn{1}{c|}{TOP80} & \multicolumn{1}{c|}{CE} & WE & \multicolumn{1}{c|}{AED} & \multicolumn{1}{c|}{TOP80} & \multicolumn{1}{c|}{CE} & WE \\ \hline
Our (2048) & \multicolumn{1}{c|}{0.08} & \multicolumn{1}{c|}{0.04} & \multicolumn{1}{c|}{0.84} & 1.13 & \multicolumn{1}{c|}{0.06} & \multicolumn{1}{c|}{0.03} & \multicolumn{1}{c|}{0.87} & 1.06 \\
Our (3072) & \multicolumn{1}{c|}{0.07} & \multicolumn{1}{c|}{0.04} & \multicolumn{1}{c|}{\textbf{0.86}} & 1.13 & \multicolumn{1}{c|}{0.06} & \multicolumn{1}{c|}{0.02} & \multicolumn{1}{c|}{\textbf{0.88}} & 1.06 \\
Our (4096) & \multicolumn{1}{c|}{0.08} & \multicolumn{1}{c|}{0.04} & \multicolumn{1}{c|}{0.83} & 1.18 & \multicolumn{1}{c|}{0.06} & \multicolumn{1}{c|}{0.03} & \multicolumn{1}{c|}{0.86} & 1.06 \\ \hline
FredsDeskew & \multicolumn{1}{c|}{10.82} & \multicolumn{1}{c|}{0.09} & \multicolumn{1}{c|}{0.54} & 109 & \multicolumn{1}{c|}{12.44} & \multicolumn{1}{c|}{0.1} & \multicolumn{1}{c|}{0.51} & 110 \\
PypiDeskew & \multicolumn{1}{c|}{16.59} & \multicolumn{1}{c|}{0.24} & \multicolumn{1}{c|}{0.2} & 141 & \multicolumn{1}{c|}{21.79} & \multicolumn{1}{c|}{2.51} & \multicolumn{1}{c|}{0.14} & 179 \\
Koo, Hyung Il \textbackslash{}etal & \multicolumn{1}{c|}{0.22} & \multicolumn{1}{c|}{0.09} & \multicolumn{1}{c|}{0.48} & 9.43 & \multicolumn{1}{c|}{0.24} & \multicolumn{1}{c|}{0.09} & \multicolumn{1}{c|}{0.48} & 82.37 \\
CMC-MSU & \multicolumn{1}{c|}{0.27} & \multicolumn{1}{c|}{0.11} & \multicolumn{1}{c|}{0.43} & 23.2 & \multicolumn{1}{c|}{-} & \multicolumn{1}{c|}{-} & \multicolumn{1}{c|}{-} & - \\
LRDE-EPITA-a & \multicolumn{1}{c|}{0.14} & \multicolumn{1}{c|}{0.06} & \multicolumn{1}{c|}{0.66} & 10.61 & \multicolumn{1}{c|}{-} & \multicolumn{1}{c|}{-} & \multicolumn{1}{c|}{-} & - \\ \hline
\end{tabular}
}
\end{table}

\noindent
\textbf{The speed of our method}. The running time of our method is around 1 second per image in the image height of $1024, 1500, 2048$ on the single-threaded implementation.
The multi-threaded implementation in our machine \footnote[1]{https://aws.amazon.com/ec2/instance-types/c5/} brings the throughput up to nearly 37 images per second in image height 1024.
Stahlberg et al. \cite{stahlberg2015document} system has got the best throughput of $5.3$ documents per minute. The LRDE-EPITA-a solution \cite{fabrizio2014precise} can process an image in around 7 seconds, our method beats this solution in both terms of accuracy and speed.


\section{Summary}
\label{sec:summary}
We have presented a novel method, which involves the Fourier transform and an Adaptive Radial Projection, to robustly solve the skew estimation problem in document image.
Our method can extract the dominant oblique angle precisely, and independently from different languages, document types and structures. 
We have also created a new high quality dataset to benchmark the skew estimators.
Finally, various experiments and detailed analyses have been contributed to the literature. 
Future works might include speed enhancements toward a real-time estimator and the evaluation of other system performance such as table extraction, OCR system, and information extraction when integrating this skew estimation method into their document processing pipeline.

\begin{center}
\section*{\small ACKNOWLDEGEMENT}
\end{center}
This research was financial, time, and facilities supported by Cinnamon AI.\\
We acknowledge the support of time and facilities from Ho Chi Minh City University of Technology (HCMUT), VNU-HCM for this study.


\label{sec:refs}

\end{document}